\documentclass[10pt,twocolumn,letterpaper]{article}

\usepackage{iccv}      

\definecolor{iccvblue}{rgb}{0.21,0.49,0.74}
\usepackage[pagebackref,breaklinks,colorlinks,allcolors=iccvblue]{hyperref}
\usepackage{xcolor}
\usepackage{tcolorbox}
\usepackage{lipsum} 
\usepackage{url}
\usepackage{tabularx}
\usepackage{booktabs}
\usepackage{array}
\usepackage[font=small,labelfont=bf,tableposition=top]{caption}
\usepackage{tcolorbox}
\usepackage{soul,colortbl}
\usepackage{float}  
\usepackage[accsupp]{axessibility}  

\usepackage{listings}
\NewDocumentCommand{\heng}
{ mO{} }{\textcolor{red}{\textsuperscript{\textit{Heng}}\textsf{\textbf{\small[#1]}}}}

\def\paperID{13260} 
\def\confName{ICCV}
\def\confYear{2025}

\title{Verbalized Representation Learning for Interpretable Few-Shot Generalization}

\author{
Cheng-Fu Yang\textsuperscript{1*} \:\: 
Da Yin\textsuperscript{1*} \:\: 
Wenbo Hu\textsuperscript{1} \:\:
Heng Ji\textsuperscript{2} \:\:
Nanyun Peng\textsuperscript{1} \:\: 
Bolei Zhou\textsuperscript{1} \:\: 
Kai-Wei Chang\textsuperscript{1} \\
\textsuperscript{1}University of California, Los Angeles \:\:
\textsuperscript{2}University of Illinois Urbana-Champaign
\\
{\tt\small cfyang@cs.ucla.edu}
}

\begin{document}


\twocolumn[{%
\renewcommand\twocolumn[1][]{#1}%

\maketitle
\begin{center}
    \centering
    \vspace{-1em}
    \captionsetup{type=figure}
    \includegraphics[page=1, trim={0 700 0 0}, clip, width=\textwidth]{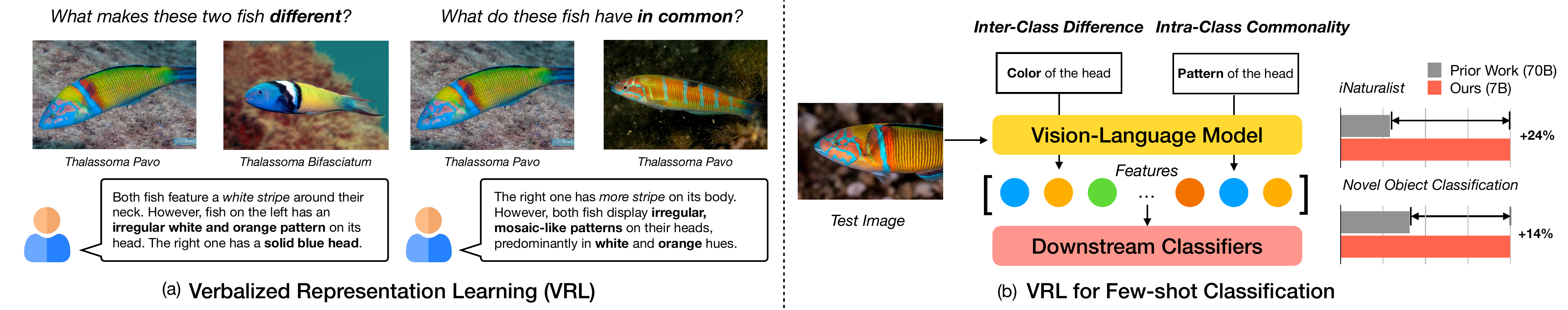}
    \captionof{figure}{(a) Humans can identify key differences between similar species and recognize common traits within a class, even when appearance varies, and express these insights in concise language. Similarly, our Verbalized Representation Learning (VRL) extracts meaningful features by querying Vision-Language Models (VLMs) to capture inter-class discriminative features and intra-class commonalities. (b) These verbalized features are mapped to numerical vectors through the VLM and can be used by downstream classifiers. Our VRL demonstrates superior performance in few-shot classification compared to prior work, achieving a 24$\%$ absolute gain on iNaturalist and a 14$\%$ on Novel Object Classification, even with a much smaller model.}
    \label{fig:teaser}
\end{center}
}]

\begin{abstract}
Humans recognize objects after observing only a few examples, a remarkable capability enabled by their inherent language understanding of the real-world environment. Developing verbalized and interpretable representation can significantly improve model generalization in low-data settings. 
In this work, we propose Verbalized Representation Learning (VRL), a novel approach for automatically extracting human-interpretable features for object recognition using few-shot data. Our method uniquely captures inter-class differences and intra-class commonalities in the form of natural language by employing a Vision-Language Model (VLM) to identify key discriminative features between different classes and shared characteristics within the same class. These verbalized features are then mapped to numeric vectors through the VLM. The resulting feature vectors can be further utilized to train and infer with downstream classifiers. Experimental results show that, at the same model scale, VRL achieves a $24\%$ absolute improvement over prior state-of-the-art methods while using $95\%$ less data and a smaller model. Furthermore, compared to human-labeled attributes, the features learned by VRL exhibit a $20\%$ absolute gain when used for downstream classification tasks. 
\footnote{The code is available at: \href{https://github.com/joeyy5588/VRL}{https://github.com/joeyy5588/VRL}.}
\vspace{-4mm}

\end{abstract}
\section{Introduction}
\label{sec:intro}


Humans have a remarkable capability to recognize certain objects after seeing only a few examples. As suggested by \cite{navarrete2020language}, it is significantly enhanced by inherent language understanding. Language, with its representational strength, serves as a primary resource for conveying knowledge about visual objects. A single, precise description can effectively capture visual distinctions observed across different object categories. For example, in Fig.~\ref{fig:teaser} (a),  a human can identify the key difference between two fish species based on the pattern of their heads, even if both share a white stripe around the neck and a yellow body. As a result, we argue that developing verbalized features offers a valuable complement to improve model's generalization under low-resource conditions. In addition, incorporating language into image classification models also increases the interpretability of the visual system, facilitating transparent and reliable decision making and effective model auditing.

Prior works~\cite{ferrari2007learning,farhadi2009describing} have explored integrating language into visual classifiers with a bottleneck of textual attributes, where the model first predicts relevant attributes and subsequently uses them to classify the image. However, these attributes often require human annotations or predefined ontologies.
Development of Large-scale Vision-and-Language Models (VLMs) like CLIP~\cite{radford2021learning} and GPT-4V~\cite{achiam2023gpt} have enabled recent bottleneck methods~\cite{menon2022visual, yan2023learning, pratt2023does, chiquier2024evolving, yang2023language} to generate relevant attributes by leveraging the prior knowledge of VLMs learned during the pre-training stage. However, these attributes are often ungrounded for the testing images and rely on VLM's prior knowledge, resulting in low precision when applied to fine-grained or novel concept recognition where objects are underrepresented in the pre-training datasets.

To address the aforementioned challenges, we propose \textit{Verbalized Representation Learning} (VRL) for \textbf{automatic}, \textbf{human-interpretable} feature extraction using only \textbf{few-shot} data. It applies to \textit{fine-grained} and \textit{novel} objects and could work with local models such as LLaVA~\cite{li2024llava}. Specifically, inspired by self-supervised representation learning (SSL) in computer vision, such as SimCLR~\cite{chen2020simple}, MoCo~\cite{he2020momentum}, SwAV~\cite{caron2020unsupervised}, SimSiam~\cite{chen2021exploring}, BYOL~\cite{grill2020bootstrap}, CLIP-Event~\cite{clipevent2022}, and SOLO~\cite{singletransformer2024}, we propose to leverage a VLM to capture the~\textit{inter-class difference} and~\textit{intra-class commonality} and articulate these findings in natural language, as illustrated in Fig.~\ref{fig:teaser}. Specifically, we cast the VLM to describe the key difference between two images from different classes, which would preserve the discriminative features and remove the redundant ones, such as the yellow body shared by both species on the left side of Fig.~\ref{fig:teaser} (a). Conversely, we employ the VLM to list the key features that are shared by two images within the same class, extracting features that are robust to intra-class variance, such as the orange facial pattern observed in all of the~\textit{Thalassoma Pavo}. Notably, this process can be applied to any two images, whether from different classes or the same class, which allows us to exponentially scale the number of useful verbalized features with the image samples, enabling effective generalization even in few-shot settings.

Consequently, to obtain the numerical feature embedding of an image using our VRL, we employ a VLM to determine whether the image possesses the characteristics described by the verbalized features, as depicted in Fig.~\ref{fig:teaser} (b). Each dimension in the resulting embedding represents a scalar value indicating the presence of the described feature. This approach effectively transforms the verbalized features into numeric representations that can serve as inputs to any classification methods, such as logistic regression, random forest, or MLP classifiers, allowing flexible and robust modeling based on the extracted interpretable features.

We conduct experiments on iNaturalist~\cite{van2018inaturalist} and Kiki-Bouba dataset~\cite{alper2024kiki}, where the former includes objects that only have subtle differences between classes, while the latter contains novel objects that are nearly not present in the web-scale datasets. Compared to previous state-of-the-art (SoTA) baselines using 70B models, we achieve a \textbf{24$\%$} absolute improvement while using \textbf{95$\%$} less data and a much smaller model with 7B parameters. Against supervised fine-tuned baselines, we observe a \textbf{15$\%$} absolute improvement. Lastly, when compared to human annotated attributes, the attributes learned by our VRL demonstrate a \textbf{20$\%$} absolute gain. These results illustrate that the features extracted by VRL not only offer superior effectiveness but also exhibit strong robustness, providing a generalizable solution when adapting to fine-grained or novel classification tasks with limited data.

\vspace{-2mm}



\section{Related Work}
\label{sec:related}

\noindent\textbf{Natural Language for Image Classification.}
A common approach for integrating language into visual classifiers involves creating a concept bottleneck~\cite{koh2020concept}. Early bottleneck methods have been extensively applied in few-shot or zero-shot classification models~\cite{ferrari2007learning,farhadi2009describing,koh2020concept, huang2016part, lampert2013attribute, frome2013devise, romera2015embarrassingly, chen2019looks}. However, these methods typically rely on manually annotated or predefined attributes, which can limit their adaptability to novel classes or fine-grained tasks. Recent advancements in VLMs have enabled researchers to directly sample interpretable descriptive features from these models~\cite{menon2022visual, yan2023learning, yang2023language, chiquier2024evolving, pratt2023does, yang2022paraphrasing, shang2024incremental, yuksekgonul2022post}. This approach benefits from incorporating external knowledge embedded in the pre-trained datasets. However, it often relies heavily on the prior knowledge encoded within these models, which can lead to the generation of ungrounded features and difficulty generalizing when the target data is not well-represented in the training datasets. In contrast, our method directly learns grounded, verbalized features from the visual data, allowing for automatic interpretable feature extraction that remains adaptable even for novel classes.

\vspace{.2em}

\noindent\textbf{Interpreting Model's Decision Process.}
There has been a long line of research aiming to improve the interpretability and explainability of deep models. Pioneering methods~\cite{selvaraju2017grad, bau2017network, bau2020understanding, zeiler2014visualizing, karpathy2015visualizing, dalvi2019one} have tackled this challenge by visualizing learned features, categorizing maximally-activating inputs, and identifying key neurons that drive model decisions. More recently, efforts have been made to interpret model behavior using natural language~\cite{hendricks2018grounding, park2018multimodal, hernandez2021natural, shaham2024multimodal}, enabling explanations that are more accessible and editable. However, the aforementioned methods are generally post-hoc, meaning they are applied after the model has been trained with abundant data. While post-hoc explanations can provide valuable insights, they may lack the ability to influence or shape the model's internal representations during training. Instead, our methods embed interpretability directly into the learning process by learning verbalized features. This enables the model to produce inherently explainable and contextually grounded representations without relying solely on retrospective analysis.
\begin{figure*}[tp]
    \centering
    \vspace{-1em}
    \scalebox{0.95}{
    \includegraphics[page=1, trim={125 325 125 0}, clip, width=\textwidth]{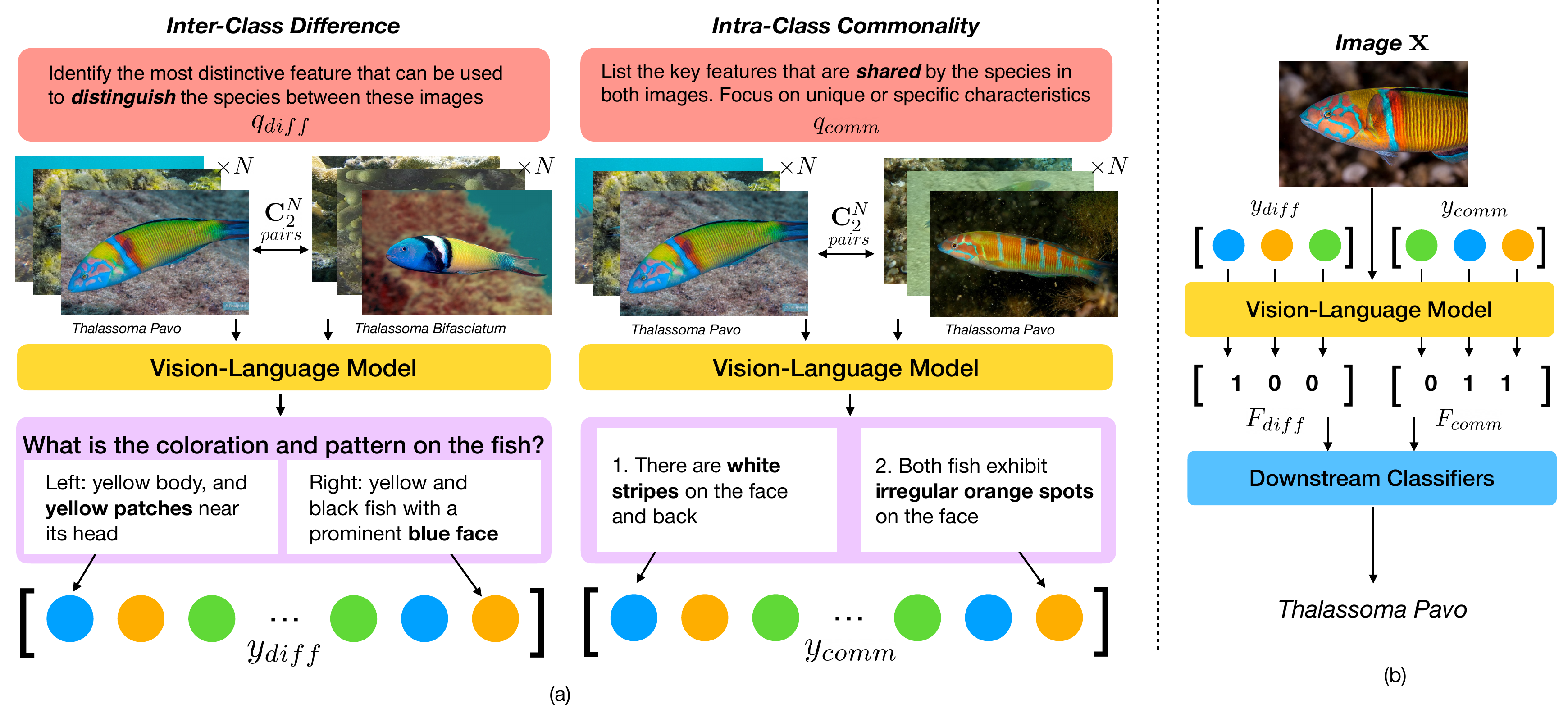}
    }
    \caption{The overview of our Verbalized Representation Learning (VRL) framework. (a) Given $N$ samples per class from $C$ different classes, VRL is able to generate a diverse set (exponentially scaling with $N \times C$) of verbalized features by: 1) extracting key differences between samples from different classes, and 2) identifying commonalities shared among objects within the same class. (b) Given an image, a Vision-and-Language model (VLM) is employed to evaluate whether the image contains the characteristics described by the verbalized features. This process can map a set of verbalized features into numeric representations which can then be used for downstream tasks.}
    \label{fig:architecture}
    \vspace{-4mm}
\end{figure*}

\vspace{.2em}

\noindent\textbf{Self-Supervised Learning Models.} Self-supervised learning (SSL) has emerged as a powerful approach in computer vision, enabling models to learn robust feature representations from unlabeled data. Contrastive learning \cite{oord2018representation} methods, such as SimCLR~\cite{chen2020simple} and MoCo~\cite{he2020momentum}, achieve this goal by learning to discriminate between different samples (negative pairs) and augmentations of the same sample (positive pairs). Chen et al.~\cite{chen2021exploring} have found that the key function of negative pairs is to prevent the model from learning collapsing features, where models produce constant or trivial outputs. As a result, subsequent works have relieved the need for negative samples by employing techniques like clustering~\cite{caron2018deep, caron2020unsupervised}, momentum update~\cite{grill2020bootstrap}, or stop gradient~\cite{chen2021exploring}. These methods focus on learning the underlying shared representations between the augmentations of the same sample. Our approach can be viewed as a variant of supervised contrastive learning~\cite{khosla2020supervised}, where the positive samples are drawn from different images from the same class. However, our method introduces two distinct advantages. First, it is data-efficient, as it does not require large datasets for gradient updates. Second, the expressiveness of verbalized features inherently prevents the model from collapsing to constant outputs, thereby ensuring the resulting features are meaningful and representative.

\section{Method}
\label{sec:method}

In Sec.~\ref{subsec:verbalized}, we first introduce the motivation and concept of~\textit{verbalized representation learning} (VRL) and then discuss how VRL automatically extracts interpretable, compact representations with few-shot data. Then, in Sec.~\ref{subsec:training}, we outline the process of building a visual classifier using the derived features. Furthermore, we demonstrate that these extracted features are versatile and can be applied to arbitrary classification models, including but not limited to logistic regression, decision trees, and multi-layer perceptron (MLP) classifiers. The overall framework of our VRL is presented in Fig.~\ref{fig:architecture}.

\subsection{Verbalized Representation Learning}
\label{subsec:verbalized}
Humans can recognize objects after seeing only a few examples, a skill boosted by language understanding \cite{navarrete2020language}. Language effectively conveys visual distinctions, even without many extra visual cues. Incorporating language also enhances image classification interpretability, enabling clearer, more reliable decisions. To this end, we propose a method that verbalizes key visual features in natural language. In this section, we focus on identifying which visual features are crucial for few-shot image classification and how to extract and verbalize them effectively, and how to map these verbalized features into vectors which would be later utilized during training and inference.
\vspace{-4mm}
\paragraph{What visual features should be verbalized?} Inspired by self-supervised representation learning methods, where model learns robust features through contrasting positive and negative pairs~\cite{chen2020simple, he2020momentum} or maximizing the similarity of augmented views of the positive samples~\cite{caron2020unsupervised, grill2020bootstrap, chen2021exploring}, we propose to verbalize the objective functions of these self-supervised learning methods with the help of VLMs such as LLaVA. Specifically, given a classification task with $C$ classes and $N$-shot examples per class, each data point is defined as $(x, l)$, where $x$ represents an image, and $l \in \{1, 2, \ldots, C\}$ is the categorical label indicating the class to which the image belongs. Our method leverages two types of paired images: positive pairs, defined as image pairs $(x_i, x_j)$ with the same label ($l_i = l_j$), and negative pairs, which consist of images $(x_i, x_k)$ from different classes ($l_i \neq l_k$). Notably, with this pairing strategy, we can form $\mathbf{C}^C_2 \times \mathbf{C}^N_2$ distinct negative pairs, and $C \times \mathbf{C}^N_2$ positive pairs, which significantly increase the data utilization under the few-shot setting.

Inspired by contrastive learning methods~\cite{chen2020simple, he2020momentum}, for each negative pair sample, our approach emphasizes capturing \textit{inter-class differences} by tasking the VLM to describe key distinguishing features between images from different classes, denoted as $y_{diff} = \text{VLM}(\mathbf{x}_i, \mathbf{x}_k, q_{diff})$, where $l_i \neq l_k$
and $q_{diff}$ denotes the query that captures the visual differences between the two input images $\mathbf{x}_i$ and $\mathbf{x}_k$. For instance, in Fig.~\ref{fig:architecture} (a), the model learns to distinguish two fish species by the coloration and the pattern of the fish. Conversely, for each positive pair, we draw inspirations from negative sample-free SSL methods~\cite{caron2020unsupervised, grill2020bootstrap, chen2021exploring}, where we employ the VLM to capture~\textit{intra-class commonalities}, generating descriptions of the key shared features between two images from the same class, i.e., $y_{comm} = \text{VLM}(\mathbf{x}_i, \mathbf{x}_j, q_{comm})$, where $l_i = l_j$ 
and $q_{comm}$ denotes the query that captures the visual commonalities between $\mathbf{x}_i$ and $\mathbf{x}_j$. For example, in Fig.~\ref{fig:architecture} (a), VRL identifies that both fish possess irregular orange spots around their face. We include the detailed prompt templates used in the above process in Appendix~\ref{subsec:template}.

Intuitively, $y_{diff}$ and $y_{comm}$ learn complimentary features. $y_{diff}$ captures the most discriminative features while filtering out redundant ones, such as the yellow body present in both types of fish, leading to less noisy features. While $y_{comm}$ is robust to the intra-class variance where it learns to identify the fish based on the shared features, like the pattern on its face, rather than the number of stripes on its back. In the later experimental section, we will further verify this assumption and show that these two features can yield the best performance when combined. 
In addition, as discussed in the earlier paragraph, even with only $N$-shot samples, our VRL can sample a diverse set of features from $\mathbf{C}^C_2 \times \mathbf{C}^N_2$ negative and $C \times \mathbf{C}^N_2$ positive pairs, making our method excel under low-resource setting. Moreover, we empirically discover that sampling $C \times N$ pairs for both negative and positive samples is sufficient to collect a robust set of features.
\vspace{-6mm}
\paragraph{How to map the verbalized features to numeric feature embeddings?} 

To obtain feature vectors that can be later used to build visual classifiers, we transform the verbalized features into numeric vectors with the help of VLMs such as CLIP~\cite{radford2021learning} and LLaVA~\cite{li2024llava}. Given an image and a set of verbalized features, which are in the form of language descriptions, we employ a VLM to determine whether the image possesses those features, as illustrated in Fig.~\ref{fig:architecture} (b), resulting in a feature embedding $F$, i.e., $F = \text{VLM}(y, x)$. Each dimension of $F$ is produced by the VLM, indicating the presence of a certain verbalized feature from $y_{diff}$ and $y_{comm}$ or assessing the degree that the image has a certain feature. Concretely, for generative VLMs like LLaVA, each verbalized feature is mapped to 0 or 1, based on whether the model infers the presence of the feature in the image. For VLMs like CLIP, feature embeddings are derived by calculating the similarity between each verbalized feature and the image using CLIP’s visual encoders. This results in a continuous vector indicating the likelihood that the image contains each feature, with an optional similarity threshold to convert it into binary feature vectors.

\subsection{Training and Inference}
\label{subsec:training}

\paragraph{Training.} Once the numeric feature vectors are generated, they are used as input for training various visual classifiers, including, but not limited to, logistic regression, random forests, Naive Bayes, K-nearest neighbors (KNN), and MLP classifiers. These models learn to predict the class labels based on the interpretable feature vectors. This flexibility enables robust modeling that can be tailored to different applications. Moreover, this approach not only enables the construction of classifiers for novel concepts with few-shot data, but also provides insight into which visual features are most important for decision-making. For instance, we can extract feature importance from logistic regression or visualize decision paths in decision tree classifiers, enhancing the interpretability of our method.

\vspace{-5mm}

\paragraph{Inference.} During inference, given a testing image, we generate its feature vector by following the approach described in Sec.~\ref{subsec:verbalized}. Specifically, we query the VLMs with the learned verbalized features to determine whether the image contains specific features, producing a numeric feature vector. This vector can either be continuous, representing the likelihood of a feature's presence, or binary, indicating the feature's absence or presence. The resulting vector is then passed to the trained classifier to make a prediction. Notably, to enhance the model robustness, we ensemble results from classifiers trained with different algorithms. Combining predictions from multiple classifiers is beneficial because it reduces the likelihood of overfitting to any specific algorithm's biases or weaknesses. The ensemble can use either hard or soft voting: In hard voting, each classifier makes a discrete prediction, and the final prediction is determined by a majority vote. In soft voting, the prediction logits from each classifier are averaged to produce the final output, allowing for more nuanced decision-making.

\section{Experiments}
\label{sec:experiment}
Our experiments are designed to answer the following research questions: (i) How well does VRL generalize to tasks requiring fine-grained recognition with few-shot data? (ii) Can VRL effectively adapt to tasks involving novel concepts that were not part of the VLM's pre-training resources, using only few-shot data? (iii) How effective is VRL compared to conventional few-shot adaptation algorithms like LoRA fine-tuning or in-context learning? (iv) Do different types of features extracted by VRL mutually benefit each other, and do the extracted features further improve on top of the other commonly used features? (v) How does the performance of automatically features extracted with VRL compare to human-labeled features?
\vspace{-1mm}
\subsection{Dataset and Evaluation Protocols}
\label{subsec:dataset}
\vspace{-2mm}

We conduct experiments on iNaturalist and Kiki-Bouba to validate our method's effectiveness under varied scenarios, with all reported numbers representing classification accuracy (\%). Although our primary analysis centers on these datasets, additional results for general few-shot classification on mini-ImageNet~\cite{ravi2017optimization} are provided in the Appendix~\ref{subsec: miniimagenet}.

\vspace{-4mm}
\paragraph{Fine-Grained Species Classification.} For fine-grained classification, we utilize the iNaturalist 2021 dataset~\cite{van2018inaturalist}, which comprises a diverse collection of images and annotations contributed by citizen scientists, spanning numerous species of animals, plants, and fungi. Each species in the training split contains between 200 and 300 images, while the validation split includes 10 images per species.

Following~\cite{chiquier2024evolving}, we experiment on images from five different families, each containing five to six species. These families are selected due to their challenging nature: distinguishing between species within the same family requires the model to identify complex features such as shapes and patterns, rather than simple color variations. The families used in our experiments are as follows: \textit{Lichen} (fungi), \textit{Wrasse} (fish), \textit{Wild rye} (grass), \textit{Manzanita} (berry shrubs), and \textit{Bulrush} (herbs). For a detailed list of the species, please refer to the appendix. Notably, unlike~\cite{chiquier2024evolving}, which utilizes a full training set containing between 200 and 300 images per species, our method uses on only~\textbf{10} images per species to achieve generalization in a low-resource setting. We will report the baseline performance from their original paper as well as our reproduced results using the same limited data as our method for comparison.
\vspace{-4mm}

\paragraph{Novel Concept Classification.} To test the model's generalizability on objects that have been rarely seen in the pre-training image resources, we evaluate our method using the Kiki-Bouba~\cite{alper2024kiki} dataset. The Kiki-Bouba experiment, originally introduced by~\cite{ramachandran2001synaesthesia}, illustrates that people often associate specific shapes with different sounds. The dataset was constructed by prompting generative models trained to create 3D-rendered images from non-existent, meaningless words~\cite{alper2024kiki}, providing a unique testbed for evaluating generalization to novel and abstract concepts. Following the approach in~\cite{chiquier2024evolving}, we evaluate our method on two distinct splits, each containing five different classes. The first split consists of the classes \textit{bamaba, duludu, gaduga, lomulo, nomano}, while the second split includes \textit{bouba, galaga, kepike, kiki, maluma}. In~\cite{chiquier2024evolving}, the training set comprises 800 images per class, with 200 images per class for validation. In contrast, we adopt a similar low-resource setting as used for the iNaturalist dataset, with our method accessing only~\textbf{10} training images per class. Baseline results for both settings will be reported in the table for comparison.
\vspace{-1mm}
\subsection{Baseline Methods and Implementation Details}
\vspace{-1mm}

To demonstrate the effectiveness of our method, we compare it against several baselines, including the CLIP-based approach proposed in~\cite{chiquier2024evolving} and other baseline methods that utilize LLMs, such as LaBo~\cite{yang2023language}, CBD~\cite{menon2022visual} and LLM-Mutate~\cite{chiquier2024evolving}, which leverage LLM's pre-trained knowledge to generate attributes. Additionally, since our method leverages LLaVA-OneVision~\cite{li2024llava} to generate verbalized features, we perform quantitative comparisons with common approaches used to adapt LLaVA for downstream tasks, such as LoRA fine-tuning~\cite{hu2021lora} and in-context learning~\cite{dong2022survey}. We describe the details of each baseline method as follows.

\vspace{-4mm}

\paragraph{CLIP-Based Methods.} For CLIP-based baselines, we consider two variants.~\textit{CLIP Class Name} is a naive baseline where CLIP is employed to compute the similarity between the species' scientific name and the images. The best accuracy is reported as the highest value achieved among using the common name, the scientific name, or both names combined. The second variant involves \textit{Prompt Tuning} with a CLIP encoder, where the text embeddings of specific class-related tokens are optimized through gradient descent. For both baselines, we present results as reported in~\cite{chiquier2024evolving}.

\vspace{-4mm}

\paragraph{Language Bottleneck Methods.} \textit{Classification by Description}~\cite{menon2022visual},~\textit{LaBo}~\cite{yang2023language}, and \textit{LLM-Mutate}~\cite{chiquier2024evolving} generate a list of attributes by prompting LLMs like GPT~\cite{chatgpt_paper} and Llama~\cite{touvron2023llama} with the class name of the object. While this allows them to leverage the prior knowledge learned by LLMs, the generated features often lack grounding in the actual images.~\textit{LaBo} and \textit{LLM-Mutate} alleviate this issue by using submodular optimization and evolutionary algorithm, respectively, to filter out ungrounded attributes. However, these methods still rely heavily on the capabilities and learned knowledge of the pre-trained LLMs. For \textit{LaBo}, we adapt the official implementation for our datasets; for \textit{LLM-Mutate}, we not only report the results from the original paper but also evaluate the method using the same limited data as our approach, ensuring a fair comparison.
\begin{table*}[ht]
    \centering
    \vspace{-1em}
    \scalebox{0.9}{
    \begin{tabularx}{\textwidth}{l>{\centering\arraybackslash}X>{\centering\arraybackslash}X>{\centering\arraybackslash}X>{\centering\arraybackslash}X>{\centering\arraybackslash}X>{\centering\arraybackslash}X}
    \toprule
    \textbf{Method} & \textbf{Lichen} & \textbf{Wrasse} & \textbf{Wild Rye} & \textbf{Manzanita} & \textbf{Bulrush} & \textbf{Average} \\
    \midrule
    \rowcolor[gray]{0.95} \multicolumn{7}{c}{\textbf{Zero-Shot Methods}} \\
    \midrule
    CLIP Class Name~\cite{chiquier2024evolving} & 23.3 & 32.0 & 32.0 & 26.0 & 26.0 & 27.86 \\
    \midrule
    \rowcolor[gray]{0.95} \multicolumn{7}{c}{\textbf{Full Dataset Methods (200+ Images per Species)}} \\
    \midrule
    CLIP Prompt Tuning~\cite{chiquier2024evolving} & 23.3 & 20.0 & 40.0 & 20.0 & 20.0 & 24.66 \\
    Classification by Description~\cite{menon2022visual} & 30.0 & 34.0 & 36.0 & 28.0 & 20.0 & 29.60 \\
    LLM-Mutate-70B~\cite{chiquier2024evolving} (1-prompt) & 31.6 & 24.0 & 44.0 & 40.0 & 22.0 & 32.32 \\
    LLM-Mutate-70B~\cite{chiquier2024evolving} (10-prompt) & 48.3 & 44.0 & 58.0 & 58.0 & 42.0 & 50.06 \\
    \midrule
    \rowcolor[gray]{0.95}
    \multicolumn{7}{c}{\textbf{Few-Shot Methods (10 Images per Species)}} \\
    \midrule
    LLM-Mutate-7B$^{\dagger}$ (10-prompt) & 35.0 & 48.0 & 38.0 & 44.0 & 26.0 & 38.20 \\
    LLM-Mutate-70B$^{\dagger}$ (10-prompt) & 46.6 & 44.0 & 46.0 & 44.0 & 40.0 & 44.13 \\
    LaBo-7B$^{\dagger}$~\cite{yang2023language} & 45.0 & 48.0 & 44.0 & 42.0 & 36.0 & 43.00 \\
    LaBo-70B$^{\dagger}$~\cite{yang2023language} & 56.6 & 46.0 & 64.0 & 52.0 & 40.0 & 51.72 \\
    LLaVA-ICL-7B & 16.6 & 28.0 & 22.0 & 18.0 & 30.0 & 22.92 \\
    LLaVA-SFT-7B & 41.6 & 50.0 & 58.0 & 42.0 & 28.0 & 43.92 \\
    LLaVA-VRL-7B (\textbf{Ours}) & 58.3 & 48.0 & 74.0 & \textbf{66.0} & 46.0 & 58.46 \\
    LLaVA-VRL-72B (\textbf{Ours}) & \textbf{71.6} & \textbf{72.0} & \textbf{74.0} & 56.0 & \textbf{66.0} & \textbf{67.92} \\
    \bottomrule
    \end{tabularx}
    }
    \caption{Comparison of classification accuracy (\%) across different methods for fine-grained classification on iNaturalist. The table presents results for zero-shot, full-dataset (200+ images per species), and few-shot (10 images per species). Results marked with $^{\dagger}$ denote values reproduced using the official implementation but restricted to the same few-shot data as our method.}
    \label{table:finegrained}
    \vspace{-6mm}
\end{table*}

\vspace{-4mm}

\paragraph{LLaVA-Based Methods.} We include two common approaches for adopting LLaVA. The first baseline involves performing LoRA~\cite{hu2021lora} fine-tuning (\textit{LLaVA-SFT}), which has been demonstrated to be effective in scenarios with limited data. The second baseline is in-context learning~\cite{dong2022survey} (\textit{LLaVA-ICL}), where one exemplar image for each species or object is included in the prompt to guide the model's predictions. We study which of the LLaVA-based methods and VRL better utilizes LLaVA's capabilities.

\vspace{-4mm}

\paragraph{Our Method.} We implement VRL with LLaVA-OneVision for capturing visual difference and commonality features. We also select LLaVA-OneVision as the feature mapping model that assists to convert an image into a feature vector, and the base model for all the LLaVA-based baseline methods. For the few-shot training stage, we experiment with various classification methods including logistic regression or MLP. Unless otherwise specified, we utilize both the difference features $F_{diff}$ and commonality features $F_{comm}$. We report the best performance achieved across different classifiers, selecting the optimal results from a \textit{single} classifier.

\vspace{-2mm}

\subsection{Fine-Grained Species Classification}
\vspace{-1mm}
We present the results for fine-grained species classification on the iNaturalist dataset in Table~\ref{table:finegrained}. We find that compared to previous SoTA methods, even with $95\%$ less data (10 images per species compared to 200+ images per species), and using a significantly smaller model (7B v.s. 70B parameters), our method can already surpass the previous SoTA by $8\%$. When we increase the scale of our method to a comparable 72B model, we can further increase the improvements to nearly $18\%$. Moreover, when baseline methods are limited to the same data availability as our approach, our 7B and 72B models achieve improvements of $20\%$ and $23\%$, respectively. These results highlight the advantage of using grounded descriptions to capture the subtle visual cues, especially when the model needs to discover visual nuances to discriminate species within the same family. 

We also compare VRL against LLaVA-based baselines. Using the same backbone LLaVA model, our method showcases a $15\%$ gain to \textit{LLaVA-SFT} and a $36\%$ advantage over the in-context learning method, \textit{LLaVA-ICL}. In addition, VRL is easier to be scaled up than the baselines. For instance, VRL with the 72B model requires 8 GPUs with 48 GB of RAM, while fine-tuning the 13B model with LoRA already requires 8 GPUs with 80 GB of RAM. For in-context learning, including image examples for every category in the prompt significantly increases the memory usage of the method, making it infeasible to perform in-context learning on the 72B LLaVA with the 8 GPUs (48 GB). The model performance may also be constrained by the context window of LLaVA model. These results and analysis further validate the scalability and robustness of VRL comparing to those existing adaptation algorithms. 

\vspace{-1mm}
\subsection{Novel Concept Classification}
\vspace{-2mm}

For novel concept classification, we report the results on the Kiki-Bouba dataset in Table~\ref{table:kiki}. Similar to the performance trend observed in the previous experiment, our method surpasses the previous SoTA by $14\%$ with a smaller model, but with $99\%$ less data (10 images per species compared to 800+ images per species). Surprisingly, we observe that increasing the number of model parameters does not lead to better performance on this task. One possible explanation for this is that, rather than relying on prior knowledge learned during pre-training, the model needs to focus on learning new, critical features specific to these novel objects with unfamiliar shapes and patterns. 
Since the objects in this task are unseen during LLM training, existing models struggle to generate relevant attributes and fail to ground visual differences necessary for distinguishing objects from different classes.
This hypothesis is further supported by the larger performance gap between our method and those that rely on LLM's pre-trained knowledge~\cite{menon2022visual}. On this dataset, our method exceeds their performance by $50\%$, compared to a $37\%$ advantage on the previous dataset. 
\begin{table}[tp]
    \centering
    \vspace{-1em}
    \scalebox{0.9}{
    \begin{tabularx}{0.5\textwidth}{l>{\centering\arraybackslash}X>{\centering\arraybackslash}X>{\centering\arraybackslash}X}
    \toprule
    \textbf{Method} & \textbf{v1} & \textbf{v2} & \textbf{Avg.} \\
    \midrule
    \rowcolor[gray]{0.95}
    \multicolumn{4}{c}
    {\textbf{Zero-Shot Methods}} \\
    \midrule
    CLIP Class Name & 38.7 & 38.8 & 38.75 \\
    \midrule
    \rowcolor[gray]{0.95}
    \multicolumn{4}{c}{\textbf{Full Dataset Methods (800+ Images per Object)}} \\
    \midrule
    CLIP Prompt Tuning & 16.7 & 55.6 & 36.15 \\
    Classification by Description & 28.8 & 36.8 & 32.80 \\
    LLM-Mutate-70B (1-prompt) & 50.3 & 47.8 & 49.05 \\
    LLM-Mutate-70B (10-prompt) & 79.2 & 59.4 & 69.30 \\
    \midrule
    \rowcolor[gray]{0.95}
    \multicolumn{4}{c}{\textbf{Few-Shot Methods (10 Images per Object)}} \\
    \midrule
    LLM-Mutate-7B$^{\dagger}$ (10-prompt) & 63.3 & 59.2 & 61.25 \\
    LLM-Mutate-70B$^{\dagger}$ (10-prompt) & 67.3 & 59.2 & 63.25 \\
    LLaVA-ICL-7B & 24.5 & 27.2 & 25.85 \\
    LLaVA-SFT-7B & 72.1 & 50.6 & 61.35 \\
    LLaVA-VRL-7B (\textbf{Ours}) & \textbf{89.4} & \textbf{76.6} & \textbf{83.00} \\
    LLaVA-VRL-72B (\textbf{Ours}) & 89.1 & 74.7 & 81.90 \\
    \bottomrule
    \end{tabularx}
    }
    \caption{Accuracy (\%) across different methods for novel concept classification on Kiki-Kouba dataset. v1 and v2 indicates two different splits described in Sec.~\ref{subsec:dataset}. The table presents results for zero-shot, full-dataset, and few-shot. Results marked with $^{\dagger}$ denote values reproduced using the official implementation but restricted to the same few-shot data as our method.}
    \label{table:kiki}
    \vspace{-10mm}
\end{table}

\subsection{Additional Analysis}
\paragraph{Fusing Features Learned by VRL and Commonly Used Features.}
Table~\ref{table:ensemble} summarizes the classification accuracy on iNaturalist achieved by fusing different types of visual features. Specifically, we study the two feature vectors: $F_{diff}$ and $F_{comm}$ learned via our VRL, which focusing on capturing regional nuances of the object. $F_{diff}$ captures the inter-class difference while $F_{comm}$ represents the shared features within the same class. 
We also consider the image features directly encoded by the vision encoder of CLIP, which often encode high-level semantics such as the object type, context, and other conceptual associations. As discussed in Sec.~\ref{subsec:verbalized}, these features are robust to different kinds of variance and could be benefited when combined for downstream tasks. This is evident from the $2\%$ improvements when concatenating $F_{diff}$ and $F_{comm}$, and an additional $6\%$ gain after including the CLIP feature. 

\begin{table}[tp]
    \centering
    \vspace{-1em}
    \scalebox{0.85}{
    \begin{tabularx}{0.5\textwidth}{l>{\centering\arraybackslash}X>{\centering\arraybackslash}X>{\centering\arraybackslash}X>{\centering\arraybackslash}X}
    \toprule
    \textbf{Method} & \textbf{$F_{diff}$} & \textbf{$F_{comm}$} & \textbf{CLIP} & \textbf{Avg.} \\
    \midrule
    Concat & \checkmark & & & 65.26 \\
    Concat & & \checkmark & & 58.32 \\
    Concat & & & \checkmark & 63.26 \\
    Concat & \checkmark & \checkmark & & 67.92 \\
    Concat & \checkmark & \checkmark & \checkmark & 73.92 \\
    \midrule
    Ensemble (hard) & LR & LR & LR & 68.32 \\
    Ensemble (hard) & MLP & MLP & MLP & 74.26 \\
    Ensemble (hard) & LR & LR & MLP & 71.60 \\
    Ensemble (soft) & LR & LR & LR & 67.06 \\
    Ensemble (soft) & MLP & MLP & MLP & 74.92 \\
    Ensemble (soft) & LR & LR & MLP & \textbf{76.52} \\
    \bottomrule
    \end{tabularx}
    }
    \caption{Accuracy (\%) when incorporating feature vectors learned from different methods. $F_{diff}$ and $F_{comm}$ denote the difference and commonality feature vectors learned from VRL. CLIP refers to the image features encoded by CLIP visual encoder. We compare two approaches: concatenating all features to train a single classifier and ensembling classifiers trained separately on each feature type. The visual classifiers include Logistic Regression (LR) and Multi-Layer Perceptron (MLP) models.}
    \label{table:ensemble}
    \vspace{-10mm}
\end{table}
We also discover that ensemble classifiers trained separately on each feature type are more effective than simply concatenating all features to build a single classifier. This is primarily because these features are heterogeneous. For example,  $F_{diff}$ and $F_{comm}$ are generally binary, with each element indicating the presence of a specific verbalized feature. Therefore, they tend to perform best when used with logistic regression models. Meanwhile, the features encoded by CLIP are continuous, which makes them more suitable for mapping to class labels through MLP models. Empirically, we demonstrate that the best configuration of ensembling can further improves the performance by $3\%$ comparing to feature concatenation. We find that with three classifiers, hard-voting is prone to high variance, while soft-voting is more robust since it leverages the confidence of each classifier by averaging their probabilities.
These experiments showcase that our approach is flexible and can be seamlessly integrated with existing representations to build a robust classification model through ensembling.
\vspace{-4mm}
\paragraph{Comparison with Human-Labeled Features.}
To evaluate the performance of automatically extracted features compared to human-labeled features, we conduct experiments on the Kiki-Bouba dataset using the human-annotated attributes provided by~\cite{chiquier2024evolving}. From Table~\ref{table:human}, we observe that VRL extracted features outperform human-labeled attributes across all configurations. Our best configuration, which combines both difference feature vector $F_{diff}$ and commonality feature vector $F_{comm}$, surpasses the human-derived features by $20\%$. This underscores the effectiveness of our automatic feature extraction approach in capturing relevant and distinguishing characteristics for classification tasks. These results also suggest that, in scenarios with limited resources or new visual concepts, we can automate the process of feature extraction, reducing the need for manual human annotation.


\vspace{-4mm}
\begin{figure*}[tp]
    \centering
    \vspace{-1em}
    \includegraphics[page=1, trim={0 375 75 0}, clip, width=\textwidth]{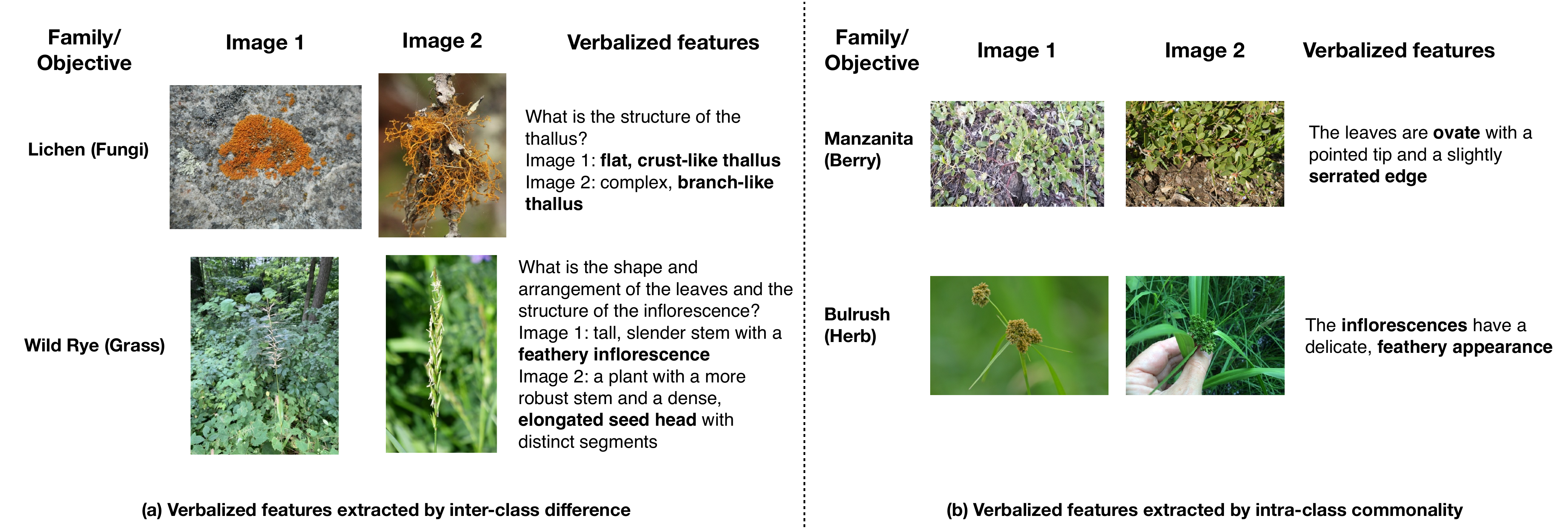}
    \caption{Qualitative examples of the features extracted by VRL. We highlight the key attributes in~\textbf{bold}. (a) Verbalized features extracted by comparing images from different classes. (b) Verbalized features extracted by comparing the images within the same class.}
    \label{fig:qualitative}
    \vspace{-4mm}
\end{figure*}

\paragraph{Classifier Choices and Feature Mapping Models.}
We investigate the model performance using different choices for visual classifiers and feature mapping models. From Table~\ref{table:inference}, we notice that LLaVA, as a feature mapping model for obtaining binary feature vectors, achieves the best overall performance when paired with the logistic regression. 

As mentioned in Sec.~\ref{subsec:verbalized}, we can also leverage CLIP as a feature mapping model to convert images into features. When predicting, one can either set a threshold to determine if the image possesses a certain attribute, resulting in a binary feature, or preserve the similarity between the image and each attribute to form continuous features. 
Although the performance of using CLIP features is slightly lower compared to LLaVA, CLIP offers a notable advantage in terms of inference speed. This is because CLIP supports batch-wise operations when computing similarities between a set of images and a set of verbalized features, which offers our method the flexibility to scale with a larger dataset or an increasing number of object classes. 
It is worth noting that using CLIP as the feature mapping model still allows our method to surpass previous state-of-the-art results by significant margins, achieving improvements of $8\%$ and $12\%$ on the iNaturalist and Kiki-Bouba datasets, respectively. This highlights the universal applicability of VRL-extracted features, regardless of the feature mapping approach used.
\begin{table}[tp]
    \centering
    \scalebox{0.9}{
    \begin{tabularx}{0.5\textwidth}{l>{\centering\arraybackslash}X>{\centering\arraybackslash}X>{\centering\arraybackslash}X}
    \toprule
    \textbf{Method} & \textbf{v1} & \textbf{v2} & \textbf{Avg.} \\
    \midrule
    Human & 73.8 & 52.5 & 63.15 \\
    VRL-$F_{diff}$ & 88.4 & 75.2 & 81.80 \\
    VRL-$F_{comm}$ & 89.2 & 74.0 & 81.60 \\
    VRL-both & \textbf{89.4} & \textbf{76.6} & \textbf{83.00} \\
    \bottomrule
    \end{tabularx}
    }
    \caption{Accuracy (\%) of using human-labeled attributes and VRL extracted features on Kiki-Bouba dataset. Note that v1 and v2 indicates two different splits described in Sec.~\ref{subsec:dataset}.}
    \label{table:human}
    \vspace{-8mm}
\end{table}
\vspace{-2mm}
\subsection{Qualitative Analysis}
Figure~\ref{fig:qualitative} presents qualitative examples of what verbalized features are extracted from image pairs. We list some of the features that have top-5 feature importance in our trained logistic regression model. For inter-class differences (Fig.~\ref{fig:qualitative} (a)), verbalized features are used to capture distinguishing characteristics between different species. For example, the Lichen (Fungi) category is distinguished based on the structure of the thallus, where one image describes a flat, crust-like thallus, and the other describes a complex, branch-like thallus. Similarly, in the Wild Rye (Grass) category, the differences between a slender stem with a feathery inflorescence and a more robust stem with a dense, elongated seed head are verbalized. For intra-class commonality (Fig.~\ref{fig:qualitative} (b)), verbalized features are employed to capture shared traits within the same species. For example, the leaves of Manzanita (Berry) are consistently described as ``ovate with a pointed tip and slightly serrated edges,'' while the inflorescences of Bulrush (Herb) are uniformly described as having a ``delicate, feathery appearance.'' These verbalized features are used to guide the decision process by helping the model focus on important distinguishing features or shared traits within each class. Due to page limit, see Appendix~\ref{subsec:hallucinate} for human study on the learned verbalized features.
\begin{table}[t]
    \centering
    \scalebox{0.8}{
    \begin{tabularx}{0.5\textwidth}{l>{\centering\arraybackslash}X>{\centering\arraybackslash}X>{\centering\arraybackslash}X}
    \toprule
    \textbf{Method} & \textbf{RF} & \textbf{LR} & \textbf{MLP} \\
    \midrule
    \rowcolor[gray]{0.95}
    \multicolumn{4}{c}{\textbf{iNaturalist (Prior Works: 50.1)}} \\
    \midrule
    LLaVA & 64.9 & \textbf{67.4} & 66.0 \\
    CLIP (binary) & 52.2 & 58.0 & 55.2 \\
    CLIP (continuous) & 52.8 & 53.6 & 60.6 \\
    \midrule
    \rowcolor[gray]{0.95}
    \multicolumn{4}{c}{\textbf{Kiki-Bouba (Prior Works: 69.3)}} \\
    \midrule
    LLaVA & 83.65 & \textbf{86.65} & 85.65 \\
    CLIP (binary) & 79.25 & 81.50 & 81.10 \\
    CLIP (continuous) & 77.05 & 77.10 & 81.85 \\
    \bottomrule
    \end{tabularx}
    }
    \caption{Accuracy (\%) of different classifier choices (columns) and models used to map verbalized features to numeric vectors (rows). 
    LLaVA and CLIP are utilized as mapping models to map an image to a feature vector.}
    \label{table:inference}
    \vspace{-9mm}
\end{table}

\vspace{-2mm}
\section{Conclusion}
In this paper, we propose Verbalized Representation Learning (VRL), which enables automatic interpretable feature extraction with few-shot samples. By leveraging VLMs, VRL generates verbalized features that capture both inter-class differences and intra-class commonalities. Our method not only enhances the model's adaptability with limited data but also provides transparency in the decision-making process, enabling easier interpretation of the features that influence predictions. Our experiments show that VRL outperforms prior approaches, achieving superior results while using significantly less data. This includes tasks like fine-grained recognition and novel concept adaptation, demonstrating its potential for real-world applications.

\clearpage

\section*{Acknowledgements}

We thank anonymous reviewers and other members of UCLA-NLP+ group for their helpful comments. This work was partially supported by U.S. DARPA ECOLE Program No. \#HR00112390060, ONR grant N00014-23-1-2780, DARPA ANSR program FA8750-23-2-0004, and Apple. The views and conclusions contained herein are those of the authors and should not be interpreted as necessarily representing the official policies, either expressed or implied, of DARPA, or the U.S. Government. The U.S. Government is authorized to reproduce and distribute reprints for governmental purposes notwithstanding any copyright annotation therein.
{
    \small
    \bibliographystyle{ieeenat_fullname}
    \bibliography{main}
}

\clearpage
\setcounter{page}{1}
\maketitlesupplementary
\setcounter{page}{1}
\setcounter{section}{0} 
\renewcommand{\thesection}{\Alph{section}} %

\section{Implementation Details}

In this section, we outline the detailed prompt template used to generate verbalized features, and the hyperparameters used throughout the experiment. Since the proposed verbalized representation learning (VRL) only involves inference using Vision Language Models, we are able to significantly improve the inference speed and the GPU memory usage by leveraging existing optimization techniques. Specifically, we utilize Sglang~\cite{zheng2023sglang}, which introduces optimizations such as RadixAttention for KV cache reuse to accelerate inference. In our experiments, we use LLaVA-OneVision as the VLM, since it is able to interleave multiple images in the prompt. For GPU usage, the 7B model requires 2 A6000 GPUs, each with 48GB of RAM, while the 72B model demands 8 A6000 GPUs to host the model.

\subsection{Prompt Templates in VRL}
\label{subsec:template}

\lstset{
    basicstyle=\ttfamily\footnotesize,
    columns=fullflexible,
    breaklines=true,
    frame=single,
    backgroundcolor=\color{gray!10},
    keywordstyle=\color{blue},
    stringstyle=\color{red},
    showstringspaces=false,
}
\begin{table}[tp]
    \centering
    \begin{lstlisting}[escapechar=|]
    {
        "role": "user",
        "content": [
            {
                "type": "image_url",
                "image_url": {
                    "url": "data:image/jpeg;base64,|\{\{image1\}\}|"
                },
                "modalities": "multi-images"
            },
            {
                "type": "image_url",
                "image_url": {
                    "url": "data:image/jpeg;base64,|\{\{image2\}\}|"
                },
                "modalities": "multi-images"
            },
            {
                "type": "text",
                "text":  q_diff/q_comm
            }
        ]
    }
    
    q_diff = "Identify the most distinctive feature that can be used to distinguish the species between image 1 and image 2."
    q_comm = "List the key features that not only shared by the species in both images but also make this species distinct from others. Focus on unique or specific characteristics, such as detailed patterns in the arrangement, textures, color variations, or specific forms of growth on surfaces. Provide each feature as a distinct bullet point, capturing the essence of what makes this species visually identifiable."
    \end{lstlisting}

    \caption{Prompt template for generating verbalized features. Note that $q_{diff}$ is the text query used to generate inter-class difference feature $y_{diff}$ and $q_{comm}$ is for intra-class commonality $y_{comm}$.}
    \label{prompt:ydiff}
\end{table}

\lstset{
    basicstyle=\ttfamily\footnotesize,
    columns=fullflexible,
    breaklines=true,
    frame=single,
    backgroundcolor=\color{gray!10},
    keywordstyle=\color{blue},
    stringstyle=\color{red},
    showstringspaces=false,
}
\begin{table*}[tp]
    \centering
    \begin{lstlisting}[escapechar=|]
    system_prompt (y_diff) = """
    I have a series of descriptions that I would like to convert into classification questions. For each description, respond in JSON format, which includes a question and provides specific labels for Class 1 and Class 2 based on the key distinguishing feature mentioned in the description.
    \nExample description: The most distinctive feature that can be used to distinguish class 1 and class 2 is the type of fungus present. class 1 has a bright yellow, fuzzy fungus with a round shape, while class 2 has bright yellow, delicate flower-like structures growing from a dark gray tree branch.
    \nExample response: {\"question\": \"What type of fungus is present?\", \"class_1\": \"bright yellow, fuzzy fungus with a round shape\", \"class_2\": \"bright yellow, delicate flower-like structures growing from a dark gray tree branch\"}
    """

    system_prompt (y_comm) = """
    I have a series of descriptions that I would like to convert into a list of structured sentences, where each item describes one specific feature of the species. For each description, response in a list format.
    \nExample description: The berry in both images exhibits several distinctive characteristics that set it apart from other berry species:\n\n- **Flower Structure**: The flowers are small, with five petals each, and they form in clusters. The petals are delicate and appear to be a soft pink or white color.\n- **Leaf Arrangement**: The leaves are arranged in an opposite or alternate pattern, with each leaf having a distinct shape that is often described as oval with a pointed tip.\n- **Leaf Texture**: The leaves have a velvety texture, which is unique to this species.\n- **Stem and Branches**: The stems and branches have small thorns or are spiny, which can be a defense mechanism against herbivores.\n- **Foliage Color**: The foliage is a vibrant green, indicating a healthy, thriving plant.\n- **Berries**: The berries are small, round, and appear to be a dark red or purple color, typical of many berry species.\n- **Growth Environment**: Both images show the plant growing in a rocky, perhaps alpine environment, which suggests it has adapted to grow in challenging conditions.\n- **Unique Shape**: The leaves and flowers have a unique shape, with the leaves having a slightly wavy edge and the flowers having a bell-shaped form.
    \nExample response: [\"Its flowers are small, with five petals each, and they form in clusters. The petals are delicate and appear to be a soft pink or white color.\",\"The leaves are arranged in an opposite or alternate pattern, with each leaf having a distinct shape that is often described as oval with a pointed tip.\",\"The leaves have a velvety texture, which is unique to this species.\",\"The stems and branches have small thorns or are spiny, which can be a defense mechanism against herbivores.\",\"The foliage is a vibrant green, indicating a healthy, thriving plant.\",\"The berries are small, round, and appear to be a dark red or purple color, typical of many berry species.\",\"The plant growing in a rocky, perhaps alpine environment, which suggests it has adapted to grow in challenging conditions.\",\"The leaves and flowers have a unique shape, with the leaves having a slightly wavy edge and the flowers having a bell-shaped form.\"]
    """

    user_prompt = f"Now, convert this description: {y_diff/y_comm}" + " Please follow the same JSON format for the response. Response:" 
    \end{lstlisting}

    \caption{Given the verbalized feature ($y_{diff}$ and $y_{comm}$), we use the VLM to convert the description into a question and the corresponding answer for each class.}
    \label{prompt:convertdiff}
\end{table*}

\lstset{
    basicstyle=\ttfamily\footnotesize,
    columns=fullflexible,
    breaklines=true,
    frame=single,
    backgroundcolor=\color{gray!10},
    keywordstyle=\color{blue},
    stringstyle=\color{red},
    showstringspaces=false,
}
\begin{table}[tp]
    \centering
    \begin{lstlisting}[escapechar=|]
    user_prompt (y_diff) = f"Given the following image, classify it based on the provided criteria:
    \nCriteria (Question): {question}
    \nClass 1: {class_1_ans}
    \nClass 2: {class_2_ans}
    \nPlease response with \"Class 1\" or \"Class 2\"

    user_prompt (y_comm) = f"Examine the given image and determine if it matches the features described by the following criteria: {question). Answer only with YES or NO."
    
    {
        "role": "user",
        "content": [
            {
                "type": "image_url",
                "image_url": {
                    "url": f"data:image/jpeg;base64,{image}"
                },
            },
            {
                "type": "text",
                "text": user_prompt,
            },
        ],
    }
    \end{lstlisting}

    \caption{Prompt template used to map verbalized feature ($y_{diff}$, $y_{comm}$) to numeric representations ($F_{diff}$, $F_{comm}$).}
    \label{prompt:mapdiff}
\end{table}

We report the prompt template used to generate verbalized features capturing inter-class difference ($y_{diff}$) and intra-class commonality ($y_{comm}$) in Table~\ref{prompt:ydiff}. Notably, since the generated descriptions often include multiple features, we utilize the same VLM again to parse the descriptions into a question and the corresponding answers, as shown in Table~\ref{prompt:convertdiff}. This approach enables us to disentangle the various features captured by the VLMs, making them easier to map to scalar vectors. Consequently, to extract the representation of a given image using the learned verbalized features, we prompt the VLM to determine whether the described features are present in the image, as discussed in Sec.~\ref{subsec:verbalized}. The prompt used for this stage is detailed in Table~\ref{prompt:mapdiff}. For inter-class difference features ($y_{diff}$), we assign a value of 0 if the model identifies the image as more similar to class 1 for the given attribute, and 1 if it is more similar to class 2. For intra-class commonality features ($y_{comm}$), we assign a value of 1 if the model responds with `Yes' and 0 if it responds with `No'.

\subsection{Time Complexity}

Given a classification task with $C$ classes and $N$-shot examples per class, we are able to construct $C^C_2 \times C^N_2$ and $C \times C^N_2$ pairs for inter-class and intra-class images, respectively. For example, with 5 classes and 10 images per class, we can sample $450$ and $100$ distinct pairs for inter-class and intra-class cases. In addition, even with the same image pairs, perform sampling during VLM's generation can also produce diverse verbalized features. Theoretically, to learn a set of verbalized features, the time complexity is $O(b \cdot C^C_2 \cdot C^N_2)$ for $y_{diff}$ and $O(b \cdot C \cdot C^N_2)$ for $y_{comm}$, where $b$ represents the number of samples generated by the VLM for the same image pair. While it may sound intimidating, empirically, we find that setting $b$ to 1 and perform verbalized representation learning on $C \times N$ pairs for both inter and intra cases are sufficient to learn a diverse robust features set. Specifically, for a task involving 5 classes with 10 images per class, this requires only 50 inferences, which can be completed in under 30 seconds. After obtaining the verbalized features, each training image must be mapped into numeric representations based on the learned features $y_{diff}$ and $y_{comm}$. Given that there are $C \times N$ images, and each image is evaluated against $C \times N$ descriptions, the computational complexity of this stage is $O(C^2 \cdot N^2)$. Empirically, for a task with 5 classes and 10 images, our approach requires less than 30 seconds on a single A6000 to extract verbalized features. In contrast, LLM-Mutate~\cite{chiquier2024evolving} based on text-based LLM sampling could take 22 hours as the generated features often lack visual grounding, resulting in slower convergence on discriminative features.

It is worth noting that to accelerate the feature mapping process, we can replace generative VLMs like LLaVA with encoder models like CLIP to perform similarity-based feature mapping, as discussed in Sec.~\ref{subsec:verbalized}. Since we can perform similarity computation in a two-dimensional batch-wise operation, where one dimension encapsulates all the images while the other contains all the verbalized features. As a result, the time complexity is reduced to $O(1)$, which finishes in seconds, albeit with a slight trade-off in performance, as demonstrated in Table~\ref{table:inference}.

\subsection{Hyperparameters}

In this subsection, we outline the specific parameters used to construct the visual classifiers. For implementation, we utilized the \texttt{scikit-learn}~\cite{pedregosa2011scikit} package. We use the default parameters for all classifiers. Since the primary results are based on logistic regression and multi-layer perceptron (MLP) classifiers, we provide the detailed parameters for these here and refer readers to the official \texttt{scikit-learn} documentation for details on other classifiers. For logistic regression, regularization was applied using the `l2' norm via the penalty parameter. The solver was `lbfgs', suitable for multiclass problems, and the regularization strength was controlled by C, set to 1.0. Optimization stopping criteria were determined by `tol' with the default value of 0.0001.

For MLP classifier, the network has a single hidden layer with 100 neurons and uses the ReLU activation function. Optimization is handled by the Adam solver with a learning rate of 0.001 and an L2 regularization term controlled by alpha=0.0001. The model trains for a maximum of 200 iterations with a batch size set automatically, which is theminimum of 200 and the number of training samples. Early stopping is disabled and the tolerance for optimization convergence is 0.0001.

\section{Additional Analysis}

\begin{table}[ht]
    \centering
    \vspace{-1em}
    \scalebox{0.9}{
    \begin{tabularx}{0.5\textwidth}{l>{\centering\arraybackslash}X>{\centering\arraybackslash}X>{\centering\arraybackslash}X>{\centering\arraybackslash}X}
    \toprule
    \textbf{$F_{diff}$} & \textbf{$F_{comm}$} & \textbf{CLIP} & \textbf{DINO} & \textbf{Avg.} \\
    \midrule
    \checkmark &  & & & 65.26 \\
    & \checkmark &  & & 58.32 \\
    & & \checkmark  & & 63.26 \\
    & & & \checkmark & 64.06 \\
    \checkmark & \checkmark & & & 67.92 \\
    & & \checkmark & \checkmark & 72.86 \\
    \checkmark & \checkmark & \checkmark & & 76.52 \\
    \checkmark & \checkmark &  & \checkmark & 76.86 \\
    \checkmark & \checkmark & \checkmark & \checkmark & 79.92 \\
    \bottomrule
    \end{tabularx}
    }
    \caption{Accuracy (\%) when incorporating feature vectors learned from different methods. $F_{diff}$ and $F_{comm}$ denote the difference and commonality feature vectors learned from VRL. CLIP and DINO refer to the image features encoded by CLIP and DINO visual encoder, respectively. All results are reported using the ensemble of the best-performing classifier combinations. Specifically, $F_{diff}$, $F_{comm}$, and DINO are using logistic regression while CLIP features are classified by MLP classifier. }
    \label{table:supp_fusion}
    \vspace{-6mm}
\end{table}
\subsection{Feature Fusion with existing visual encoders}
Intuitively, verbalized representation learning can be viewed as a fine-tuning process where we develop features specifically tailored to our few-shot data, but without requiring gradient update steps. To validate this perspective, we investigate whether the learned verbalized features can enhance the performance of pretrained visual encoders.

Table~\ref{table:supp_fusion} presents the results of combining difference ($F_{diff}$) and commonality ($F_{comm}$) features with features extracted by pre-trained visual encoders (CLIP and DINO). These features are incorporated via an ensemble approach, where each feature is used to train a separate classifier, and the final prediction is determined by averaging the prediction logits from all classifiers.

From the table, we observe that when features predict individually, the performance hovers around $60\%$, with $F_{diff}$ yielding the highest accuracy among the standalone features. When features are combined, significant performance improvements are achieved. Specifically, adding both verbalized features ($F_{diff}$ and $F_{comm}$) to CLIP features leads to a notable accuracy increase of $13.26\%$, while a similar $12.8\%$ improvement is observed when combining these features with DINO.

Finally, combining all four features--$F_{diff}$, $F_{comm}$, CLIP, and DINO--results in a peak performance of $79.92\%$. This validates the effectiveness of integrating verbalized features with pre-trained visual embeddings, demonstrating that verbalized representation learning provides complementary, task-specific refinements that significantly enhance model performance in few-shot learning scenarios.

\begin{table*}[ht]
    \centering
    \vspace{-1em}
    \scalebox{1.0}{
    \begin{tabularx}{\textwidth}{lll>
    {\centering\arraybackslash}X>{\centering\arraybackslash}X>{\centering\arraybackslash}X>{\centering\arraybackslash}X>{\centering\arraybackslash}X>
    {\centering\arraybackslash}X>
    {\centering\arraybackslash}X>{\centering\arraybackslash}X}
    \toprule
\textbf{Size} & \textbf{F.T.} & \textbf{F.M.} & \textbf{LR} & \textbf{RF} & \textbf{SVM} & \textbf{kNN} & \textbf{NB} & \textbf{DT} & \textbf{GB} & \textbf{MLP} \\
\midrule
7b & $y_{diff}$ & LLaVA & 52.73 & 49.33 & 47.27 & 45.80 & 50.20 & 40.07 & 47.33 & 51.73 \\
7b & $y_{comm}$ & LLaVA & 53.27 & 51.27 & 51.33 & 45.80 & 53.53 & 40.07 & 43.33 & 50.87 \\
7b & both & LLaVA & 62.06 & 56.00 & 51.40 & 51.20 & 52.06 & 37.80 & 39.60 & 55.80 \\
7b & $y_{diff}$ & CLIP  & 57.07 & 55.13 & 48.47 & 46.33 & 47.53 & 43.20 & 45.60 & 52.73 \\
7b & $y_{comm}$ & CLIP  & 46.53 & 50.47 & 45.53 & 43.40 & 19.33 & 38.80 & 42.00 & 56.67 \\
7b & both & CLIP  & 48.07 & 49.47 & 45.53 & 43.40 & 21.73 & 38.40 & 41.93 & 58.87 \\
\midrule
72b & $y_{diff}$ & LLaVA & 65.27 & 64.93 & 57.07 & 56.07 & 53.53 & 45.13 & 53.60 & 62.53 \\
72b & $y_{comm}$ & LLaVA & 58.33 & 58.07 & 53.60 & 51.47 & 52.20 & 38.73 & 48.47 & 58.33 \\
72b & both & LLaVA & 67.40 & 64.87 & 58.93 & 54.47 & 55.33 & 41.73 & 44.67 & 66.00 \\
72b & $y_{diff}$ & CLIP  & 61.07 & 57.13 & 54.07 & 47.00 & 53.60 & 42.60 & 46.13 & 57.20 \\
72b & $y_{comm}$ & CLIP  & 45.87 & 50.13 & 47.80 & 44.33 & 19.33 & 40.47 & 35.87 & 53.73 \\
72b & both & CLIP  & 53.67 & 52.80 & 50.40 & 44.73 & 19.33 & 42.93 & 40.93 & 60.60 \\
\bottomrule
    \end{tabularx}
    }
    \caption{Comparison of classification accuracy (\%) across different ablated methods for fine-grained classification on iNaturalist. Note that F.T. indicates the type of the verbalized features and F.M. refers to the model used to perform feature mapping. For different classifiers, LR denotes Logistic Regression, RF for Random forest, SVM for Support Vector Machine, kNN for k nearest neighbor, NB for Naive Bayes, DT for decision tree, GB for gradient boosting and MLP for multi-layer perceptron classifier.}
    \label{table:full}
\end{table*}

\subsection{Ablation Study}
We present the complete ablation study of our method in Table~\ref{table:full}, analyzing the performance across several dimensions. Specifically, we evaluate our model using two different sizes (7B and 72B), the impact of distinct verbalized features ($y_{diff}$ and $y_{comm}$), and the effect of using different feature mapping models (LLaVA or CLIP). Additionally, we examine the effectiveness of various classifiers, including logistic regression (LR), random forest (RF), support vector machine (SVM), k-nearest neighbor (kNN), naive Bayes (NB), decision tree (DT), gradient boosting (GB), and multi-layer perceptron (MLP). We observe that larger models (72B) consistently outperform smaller models (7B) across most classifiers and settings, showcasing the benefit of increased model capacity for capturing verbalized features. We also find that the inter-class difference features ($y_{diff}$) are generally more effective than commonality feature. However, we discover a consistent trend where the combined features (via concatenation) can yield the best overall performance (the `both' rows). For different feature mapping models, LLaVA outperforms CLIP in most scenarios, showcasing the advantage of using generative VLMs to determine the presence of a certain feature. In terms of classifier, we observe that logistic regression, random forest and MLP classifiers perform the best. On the other hand, we notice that decision tree is prone to overfitting on the training set since we only have few-shot samples, while Naive Bayes also struggle to perform well since the resulting representations are high-dimensional. 

\subsection{Mini-ImageNet}
\label{subsec: miniimagenet}

In the main paper, to evaluate how well the proposed method learn when the objects are not well-presented in the pre-training datasets of the VLMs, we conduct experiments on the iNaturalist and Kiki-Kouba datasets for fine-grained and novel object recognition. To complement these experiments and provide insight into its performance on a more general few-shot benchmark, we further evaluate our method on the mini-ImageNet dataset under the 5-way 1-shot setting across 1,000 testing episodes. We report the results in Table~\ref{tab:mini}. 

From the table one can see that our method is comparable to the recent baselines that adapt VLMs or LLMs for few-shot image classification. Notably, these baselines typically require data from training episodes to perform meta-training or instruction fine-tuning, while our method directly adapts during test time without further training.



\begin{table}[t]
\centering
\scalebox{0.95}{
\begin{tabularx}{0.46\textwidth}{cccc}    \toprule
    \textbf{Ours}            & \textbf{F. Liu~\cite{liu2024making}} & \textbf{M. Liu~\cite{liu2024envisioning}}           & \textbf{Chen~\cite{chen2023semantic}}        \\ \hline
94.27 $\pm$ 0.05 & 98.24 & 81.14 $\pm$ 0.15 & 72.31 $\pm$ 0.40 \\ 
    \bottomrule
    \end{tabularx}
}
\caption{
Top-1 Accuracy on mini-ImageNet under the 5-way 1-shot setting. Baseline results are sourced from the original paper.
}
\label{tab:mini}
\end{table}

\begin{table}[t]
\centering
\small\addtolength{\tabcolsep}{1pt}
\scalebox{0.9}{
\begin{tabularx}{0.52\textwidth}{lccccccc}    \toprule
      LLaVA & \textbf{10} & \textbf{50} & \textbf{200} & LLaVA & \textbf{10} & \textbf{50} & \textbf{200} \\   
     \hline
    w/ SFT  & 43.9 & 65.2 & 80.2 & w/ VRL & 62.1 & 80.5 & 86.2 \\
    \bottomrule
    \end{tabularx}
}
\caption{
Scaling LLaVA with SFT vs. VRL from 10 to 200 (full dataset) training images per class.
}
\label{table:data}
\end{table}

\subsection{Scalability Analysis}
\paragraph{Scaling training data.} We evaluate the proposed VRL framework on progressively larger training sets and summarize the results in Table~\ref{table:data}. Our method consistently outperforms standard supervised fine-tuning—even when using the full dataset (200 images per class). We attribute this advantage to the fact that, unlike the random weight initialization in SFT, verbalized queries provide a strong prior that guides the model toward learning task-relevant features. Importantly, we find that exhaustive pairwise enumeration is not required for these gains: sampling only 100 image pairs (out of 19,900 possible) still delivers substantial improvements. For feature extraction, we adopt the CLIP-based VRL variant (see L269–274), which enables efficient feature mapping without compromising performance.

\paragraph{Scaling object classes.} We evaluate scalability on a large iNaturalist subset with 200 families, each with 5 species (1000 species) and 10 images per species. For family-level classification, which is more heterogeneous, CLIP achieves 73\%, while our $F_{diff} + F_{comm}$ features reach 83\%. Combining them further improves performance to \textbf{90\%}. Similar trends are observed in the general recognition dataset miniImageNet (Table 12, Appendix), where VRL achieves \textbf{94.2\%}. It suggests that general VLM features can handle coarse, heterogeneous distinctions well, and since VRL builds on top of them, it naturally inherits this capability while providing additional gains by extracting finer, task-relevant features. Notably, in this experiment, we sample only 1 image pair per class combination, yet still observe substantial improvements.

\subsection{Human Study on Verbalized Features}
\label{subsec:hallucinate}

\begin{table}[t]
    \centering
    \vspace{-1em}
    \scalebox{0.9}{
    \begin{tabularx}{0.5\textwidth}{l>{\centering\arraybackslash}X>{\centering\arraybackslash}X>{\centering\arraybackslash}X}
    \toprule
    & \textbf{Acc.} & \textbf{Rel.} & \textbf{Rel. (\%)} 
    \\
    \midrule
    Lichen & 71.6 & 18.0 & 90 \\
    Wrasse & 72.0 & 20.0 & 100 \\
    Wilde Rye & 74.0 & 20.0 & 100 \\
    Manzanita & 56.0 & 17.0 & 85 \\
    Bulrush & 66.0 & 18.0 & 90 \\
    \midrule
    Average & 67.9 & 18.6 & 93 \\
    \bottomrule
    \end{tabularx}
    }
    \caption{Human Study of the learned verbalized features on iNaturalist. Acc. denotes testing classification accuracy, while Rel. represents human evaluation scores (maximum 20).}
    \label{table:humanstudy}
    \vspace{-6mm}
\end{table}
To verify the quality of the learned verbalized features, we sample 20 features per super class on iNaturalist, resulting in 100 features for human evaluation. We ask the testers to evaluate whether the generated features are faithful to the image and relevant to the target objects. For example, if a feature accurately describes the image but pertains only to the background (e.g., “\textit{the sky is blue}”), it receives a score of 0. We report the results in Table~\ref{table:humanstudy}. From the results one can see that around $93\%$ of the features were faithful to the image and useful for classification, with a $0.84$ correlation between this rate and final accuracy. In addition, we observe that the learned classifier tend to assign lower weights for those irrelevant or hallucinated features, thereby reducing their impact on final predictions.

\end{document}